\documentclass[sigconf]{acmart}
\AtBeginDocument{%
  }

\usepackage{multirow}
\usepackage{multicol}
\usepackage{amsmath}
\usepackage{pifont}
\newcommand{\xmark}{\ding{55}}
\newcommand{\cmark}{\ding{51}}
\begin{document}

\title{Rethinking Spatio-Temporal Anomaly Detection: A Vision for Causality-Driven Cybersecurity}

\title{Rethinking Spatio-Temporal Anomaly Detection: A Perspective for Causality-Driven Cybersecurity}

\author{Arun Vignesh Malarkkan}
\email{arun.malarkkan@asu.edu}
\affiliation{%
  \institution{Arizona State University}
  \city{Tempe}
  \state{Arizona}
  \country{USA}
}

\author{Haoyue Bai}
\email{haoyueba@asu.edu}
\affiliation{%
  \institution{Arizona State University}
  \city{Tempe}
  \state{Arizona}
  \country{USA}
}
\author{Xinyuan Wang}
\email{xwang735@asu.edu}
\affiliation{%
  \institution{Arizona State University}
  \city{Tempe}
  \state{Arizona}
  \country{USA}
}
\author{Anjali Kaushik}
\email{akaush39@asu.edu}
\affiliation{%
  \institution{Arizona State University}
  \city{Tempe}
  \state{Arizona}
  \country{USA}
}
\author{Dongjie Wang}
\email{wangdongjie@ku.edu}
\affiliation{%
  \institution{University of Kansas}
  \city{Lawrence}
  \state{Kansas}
  \country{USA}
}

\author{Yanjie Fu}
\email{yanjie.fu@asu.edu}
\affiliation{%
  \institution{Arizona State University}
  \city{Tempe}
  \state{Arizona}
  \country{USA}
}

\renewcommand{\shortauthors}{Malarkkan et al.}

\begin{abstract}
As cyber-physical systems grow increasingly interconnected and spatially distributed, ensuring their resilience against evolving cyberattacks has become a foundational challenge.
Spatio-temporal anomaly detection plays an important role in ensuring system security and operational integrity. 
However, current data-driven approaches, largely driven by black-box deep learning, face challenges in interpretability, adaptability to distribution shifts, and robustness under evolving system dynamics.
In this vision paper, we advocate for a \textit{causal learning} perspective to advance anomaly detection in spatially distributed critical infrastructures that grounds detection in structural cause-effect relationships. 
We identify and formalize three key directions: \textit{divergence-aware causal graph profiling}, \textit{multi-view causal graph fusion reasoning}, and \textit{Continual causal graph learning}, each contributing different strengths in uncovering dynamic cause-effect structures across time and space.
Drawing on real-world insights from systems such as water treatment infrastructures, we illustrate how causal models provide early warning signals through structural graph shifts and support interpretable anomaly analysis, tackling the limitations of black-box detectors.
To advance this vision, we outline the future research agenda centered on multi-modality, generative AI-driven, and scalable adaptive causal frameworks.
Our objective is to lay a new research trajectory toward scalable, adaptive, explainable, and spatially grounded anomaly detection systems. 
We hope to inspire a paradigm shift in cybersecurity research, promoting causality-driven approaches to address evolving threats in interconnected infrastructures.
\end{abstract}

\maketitle

\section{Introduction}
Modern critical infrastructures supporting essential services across industries such as healthcare, transportation, energy, and water management increasingly rely on spatially distributed cyber-physical systems (CPS) for uninterrupted service.
Unlike conventional IoT or enterprise systems, CPSs are exposed to both environmental noise and intentional, adversarial cyber threats, which can trigger rapid, cascading failures across interconnected infrastructure. 
This adversarial and interdependent nature makes timely, explainable, and actionable anomaly detection not just desirable but essential for public safety and operational resilience.
These infrastructures are instrumented with dense networks of sensors and actuators, continuously generating high-dimensional spatio-temporal data encompassing metrics such as performance, resource utilization, sensor readings, and network activity.
While real-time data stream provides a rich substrate for anomaly detection, it also presents major challenges due to dynamic dependencies, non-stationarity, and discerning malicious perturbations from benign fluctuations.

In response to this complexity, researchers have increasingly turned to deep learning methods for anomaly detection in multivariate time-series data, converging around three primary categories: \textit{forecasting-based models}, which predict future time-window values and flag significant deviations~\cite{TEALAB2018334, Bontemps_2016, 8836638, wu2022timesnet}, \textit{reconstruction-based models}, which learns the normal state system behavior and detect anomalies with reconstruction error~\cite{DBLP:conf/iclr/ZongSMCLCC18, 10.1609/aaai.v33i01.33011409, 10.1145/3447548.3467174, xu2021anomaly}, and \textit{graph-based models}, which exploit complex spatio-temporal correlations using graph neural networks (GNN) to detect anomalies \cite{Zhao_2020, Deng_Hooi_2021, pmlr-v162-chen22x} models. Recent studies~\cite{sarfraz2024position} also show that even simple statistical baselines (e.g., PCA) can be competitive, raising important questions about what deep and black-box models are truly adding, especially in critical domains demanding more than just raw detection accuracy.

Despite their empirical success, these deep learning models suffer from key limitations that hinder their deployment in mission-critical infrastructure. Forecasting and reconstruction-based methods are inherently opaque, offering \textbf{limited interpretability} into the underlying causes of anomalies. Graph-based models capture spatial dependencies more explicitly, but \textbf{remain sensitive to distributional drift} and often require costly retraining to adapt to system evolution. Additionally, relying solely on statistical correlation makes them susceptible to confounding and superficial fluctuations, resulting in increased false alarms and spurious alerts.

\begin{figure}[h]
  \centering
  \includegraphics[height=5cm,width=0.75\columnwidth]{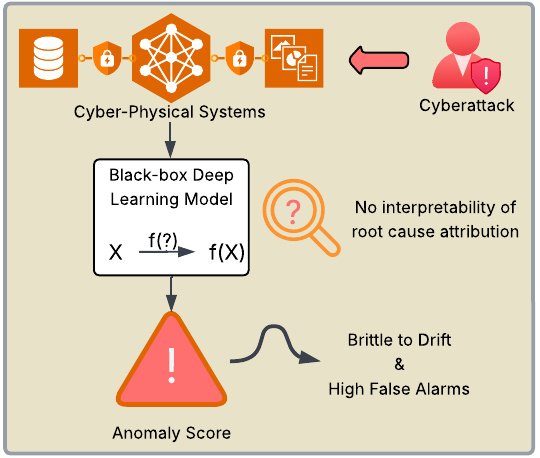}
  \vspace{-5pt}
  \caption{\small{Limitations in Current Real-Time Detection of Cyberattacks in Industrial Cyber-Physical Systems}}
  \label{limitation}
\end{figure}

Causal graph learning offers a promising principled alternative.
Grounded in Structural Causal Models (SCMs), it enables both interpretability and counterfactual reasoning by encoding how interventions (e.g., actuator changes, attack injections) propagate through the system.
Recent works have demonstrated its potential for early anomaly detection, root-cause attribution, and robust system diagnostics in cyber-physical systems~\cite{zheng2024multimodal, yokoyama2024causaldiscoverybased, 10.1145/3580305.3599392,wang2023interdependent}. 

In this vision paper, we advocate for a causality-centric paradigm for spatio-temporal anomaly detection. 
We explore the characteristics of causal graph learning models and organize our perspective around three emerging directions, beginning with \textbf{divergence-aware causal graph profiling} that focuses on causal graph representation of the system state with the multi-variate temporal data and flags structural deviations, progressing to \textbf{multi-view causal graph fusion reasoning} that seeks to integrate prior knowledge for more holistic reasoning of modeling causality in the spatio-temporal space, and finally addressing the challenges and benefits of \textbf{continual causal graph learning} which adaptively updates causal representations in response to system drift, environmental change, or adversarial tampering. These perspectives address the key challenges, including interpretability, spatial generalization, and temporal adaptability, together offering a structured foundation for building robust next-generation anomaly-detection systems. 

Looking ahead, we identify three promising research directions that extend the frontier of causality-driven anomaly detection:
1)\textbf{Multi-Modality Causal Graph Learning:} Integrating multiple data sources (e.g., sensor telemetry, system logs) to build rich, context-aware causal models;
2)\textbf{LLM-based Anomaly Detection:} 
 Leveraging Large Language Models (LLMs) to interpret, contextualize, and explain anomalous patterns in complex CPS environments.
 3)\textbf{Streaming spatio-temporal causal graph learning:} 
Enabling real-time, incremental updates to causal graphs in response to nonstationary, geographically distributed data streams.
These directions collectively aim to unify causality, spatial awareness, and generative reasoning into next-generation anomaly detection systems. Our broader goal is to foster a paradigm shift—from reactive, opaque detectors to proactive, interpretable, and intervention-ready frameworks, capable of securing critical infrastructure in an increasingly interconnected and adversarial world.

\vspace{-5pt}
\section{A Causal Perspective: Rethinking Anomaly Detection}
In the landscape of modern anomaly detection for complex, interconnected systems, black-box and even interpretable non-causal models frequently fall short: they cannot distinguish between correlation and causation, nor explain how or why anomalies emerge, especially under adversarial manipulation~\cite{10.5555/1642718, schlkopf2021causal}. In contrast, causal learning offers a principled foundation for both interpretability and robust reasoning, enabling models to answer “what-if” questions, reason about interventions, and localize true sources of system change for actionable cybersecurity in CPS~\cite{Reichenbach1956-REITDO-2, ramanan2020causal, pmlr-v202-ahuja23a, li2024causal}.

\textit{Causal graph-based anomaly detection} models the normal operational state as a Structural Causal Model (SCM), capturing directed, functional dependencies among variables under both observational and interventional distributions~\cite{schlkopf2021causal, Reichenbach1956-REITDO-2}. Each variable is a node in a directed acyclic graph (DAG), and edges encode invariant causal influences, estimated through constraint-based, score-based, or differentiable neural causal discovery algorithms~\cite{glymour2019review, ng2022masked}. Alternative frameworks, such as Granger causality~\cite{6413806}, are widely used for temporal dependence but are limited by linearity, bivariate scope, and lack of support for interventions, making them less suitable for high-dimensional, nonlinear, or adversarial CPS environments. In SCM-based approaches, anomalies are flagged when substantial deviations are detected, such as the appearance or disappearance of causal edges, or changes in edge strength, signaling violations of expected causal invariances. This provides an interpretable, topology-sensitive, and distributionally robust framework for anomaly detection in dynamic spatio-temporal environments~\cite{10.1145/3580305.3599392, 10.1145/3627673.3680096}.

\vspace{-8pt}
\subsection{Divergence-Aware Causal Graph Profiling}
Divergence-aware causal graph profiling focuses on discovering causal relationships within a single, homogenous data source to profile the system state.
For multivariate time-series anomaly detection in CPS, this involves analyzing causal dependencies among interconnected sensors across a time-series setting~\cite{10.5555/1642718}.
Established methodologies include constraint-based causal discovery (e.g., PC algorithm~\cite{spirtes2001causation}) using conditional independence tests, score-based models (e.g., Greedy Equivalence Search (GES)\cite{chickering2002optimal}) that maximize a scoring function like the Bayesian Information Criterion (BIC), and neural causal models\cite{sick2025interpretable} leveraging transformer architecture. These models can effectively capture the causal structure of multivariate time-series data.
The central idea is to treat anomalies not as statistical outliers, but as structural deviations in causality, transforming anomaly detection into the task of identifying changes in a learned causal graph structure. To envision this in network traffic monitoring, a score-based model like NOTEARS~\cite{pamfil2020dynotears} can model causal relationships between features such as packet arrival times, packet sizes, and the number of connections. Comparing the causal graph learned from normal traffic to that from potentially anomalous data enables the detection of network attacks or abnormal traffic patterns.
In our own work, we have applied causal graph profiling-based model to industrial sensor data, identifying interpretable early signs of equipment malfunction. By focusing on recurring causal disruptions rather than surface-level signal patterns, this method is resilient to distribution shifts and class imbalance that challenge traditional models.

\vspace{-8pt}
\subsection{Multi-View Causal Graph Fusion Reasoning}
Multi-view causal graph fusion leverages heterogeneous data to uncover cross-modal causal connections in complex real-world systems. A prevalent strategy involves constructing individual causal graphs for diverse data modalities using causal discovery algorithms, followed by their integration via graph-fusion techniques~\cite{zheng2024multi}. This approach yields a holistic understanding and reasoning of inter-source causal relationships, improving the accuracy and interpretability of causal analysis. For instance, in smart healthcare, integrating causal graphs from patient vital signs (time-series), medical records (text), and imaging sensor data reveals how medical history causally influences vital signs and imaging results, aiding in anomaly detection like unexpected disease progressions.
A burgeoning area is multi-view joint causal graph learning. Recent work~\cite{von2024identifiable} introduces a novel method that jointly models causal relationships across multiple data views. This technique, utilizing advanced machine learning, discerns causal structure while accounting for intricate interdependencies among modalities. This not only deepens the understanding of causal mechanisms but also significantly enhances the performance and interpretability of anomaly detection and related tasks in complex systems. In industrial cybersecurity, jointly analyzing time-series data from control systems, text-based SIEM logs, and image-based surveillance data facilitates a more robust anomaly detection system by uncovering cross-modal causal connections indicative of multi-component cyber-attacks \cite{10.1145/3627673.3680096}.

\vspace{-8pt}
\subsection{Continual Causal Graph Learning}
While traditional causal discovery methods operate in static, batch-mode settings, real-world cyber-physical systems (CPS) are dynamic, evolving due to environmental fluctuations, operator interventions, or adversarial activity. Consequently, static causal graphs quickly become obsolete in the face of temporal concept drift and evolving spatial dependencies. This motivates the need for Continual Causal Graph Learning (CCGL)—a paradigm shift that continuously adapts the causal structure of a system to reflect real-time changes.
CCGL envisions the causal graph not as a fixed representation, but as a dynamic entity that evolves with the system state. It seeks to update the causal model incrementally, without full retraining, by leveraging incoming data streams to track both structural and parametric shifts over time.
From a spatio-temporal perspective, CCGL enables continuous monitoring of causal interactions across distributed components, supporting proactive anomaly detection even under previously unseen scenarios. Unlike conventional black-box approaches, CCGL promises topology-aware explainability, revealing how and why a system's behavior deviates from expected norms. However, CCGL faces \textbf{key challenges}: (1) \textit{Concept Drift \& Temporal Dependency}—causal relationships may change over time, requiring models to distinguish transient fluctuations from structural shifts; (2) \textit{Scalability \& Computational Overhead}—updating causal graphs efficiently without excessive memory consumption is critical for real-time applications; (3) \textit{Catastrophic Forgetting}—preserving relevant causal knowledge while integrating new relationships remains a challenge; (4) \textit{Hidden Confounders}—newly emerging variables can influence causal structures, necessitating adaptive mechanisms.

We envision the following techniques address these challenges: (1) \textit{Temporal-Invariant and Temporal-Dependent Decomposition}  partitions causal structures to selectively update evolving relationships while retaining stable ones~\cite{10.1145/3580305.3599392, wang2023interdependent}; (2) \textit{Sparse Graph Updates via Laplacian Regularization} reduces redundant recomputation and improves stability~\cite{li2024nonlinear}; (3) \textit{Domain-Knowledge-Guided Pruning} \cite{10.1145/3627673.3680096} filters misleading causal edges using expert knowledge; (4) \textit{Causal Weight Reconciliation} reinforces frequently observed causal links while pruning low-confidence relationships~\cite{febrinanto2023entropy, von2024identifiable}.
Ultimately, CCGL offers a promising direction for building self-evolving, interpretable, and spatially-resilient anomaly detection systems capable of operating reliably in critical CPS environments.


\vspace{-5pt}
\section{Real-World Insights from Cyber-Physical Infrastructures}

To assess the practical viability of causal graph learning, we evaluate representative models on two widely-used cyber-physical testbeds: SWaT and WADI~\cite{7469060, 10.1145/3055366.3055375}, which feature spatially distributed sensors and actuators across water treatment and distribution stages.
We compare three causal learning approaches—\textbf{SMV-CGAD}~\cite{10.1145/3627673.3680096}, which encodes domain knowledge for structural pruning, and \textbf{CausAnom}~\cite{strelnikoff2023causanom}, a GNN-based causal encoder,  and \textbf{Entropy-CGAD} (Entropy Causal Graphs)~\cite{febrinanto2023entropy}, which uses transfer entropy and weighted GNNs against deep learning baselines like USAD~\cite{audibert2020usad}, VAE-LSTM~\cite{9053558}, and DAGMM~\cite{zong2018deep}. As shown in Table~\ref{tab:results}, causal methods match or exceed deep learning performance, while providing interpretable structural alerts.
\vskip -0.05 in
\begin{table}[h]
\caption{F1-scores for SWaT and WADI datasets.}
\label{tab:results}
\vskip -0.05in
\centering
\begin{small}
\begin{tabular}{lccc}
\toprule
Dataset & Model & Causal? & F1-Score \\
\midrule
\multirow{6}{*}{SWaT}
    & SMV-CGAD & \cmark & 0.75 \\
    & CausAnom & \cmark & 0.79 \\
    & Entropy-CGAD & \cmark & 0.82 \\
    & USAD & \xmark & 0.79 \\
    & VAE-LSTM & \xmark & 0.77 \\
    & DAGMM & \xmark & 0.55 \\
\midrule
\multirow{6}{*}{WADI}
    & SMV-CGAD & \cmark & 0.79 \\
    & CausAnom & \cmark & 0.35 \\
    & Entropy-CGAD & \cmark & 0.40 \\
    & USAD & \xmark & 0.25 \\
    & VAE-LSTM & \xmark & 0.23 \\
    & DAGMM & \xmark & 0.13 \\
\bottomrule
\end{tabular}
\end{small}
\vskip -0.05in
\end{table}

\vspace{0.5em}
\textbf{Perspectives from Causal Paradigms.} We further include:
(1) \textit{Divergence-aware causal graph profiling}, learning a causal graph during normal operations and detecting divergence during attacks (F1-0.84 on SWaT); 
(2) \textit{Continual-CGAD}, which updates causality online to adapt to drift (F1-0.85 on SWaT). While causal approaches perform on par or better than deep models in F1, their primary value lies in structural interpretability, drift robustness, and intervention-readiness. Incorporating human-in-the-loop or root-cause localization studies for interpretability evaluation remains an important direction for future work.
These results demonstrate the effectiveness of causal models to enable accurate and interpretable anomaly detection in spatio‑temporal CPS environments. 

\vspace{-5pt}
\section{Challenges and Research Opportunities}
Several limitations represent the next frontier for research towards the practical deployment of causal frameworks in CPS.

\vspace{-8pt}
\subsection{Scalability in High-Dimensional Systems}
Causal structure learning remains computationally intensive in high-dimensional CPS, particularly when the underlying Directed Acyclic Graph (DAG) spans hundreds of spatially distributed sensors and actuators. Classical causal learning approaches struggle to scale in real-time settings. Future advances must focus on causal sparsification guided by physical system topology (e.g., sensor locality, control-layer hierarchies), and exploit information-theoretic prioritization to focus learning on high-influence dependencies. Additionally, modular causal decomposition—partitioning global graphs into semi-independent subgraphs based on physical zones, can offer both tractability and local interpretability~\cite{murphy2024bits}.

\vspace{-8pt}
\subsection{Beyond Causal Sufficiency in Incomplete Spatial Systems}
The assumption of \textit{causal sufficiency}—that all confounders are observed, is often violated, as sensor failures, communication lags, or limited coverage can lead to missing causal sources. In these cases, causal discovery algorithms risk inferring spurious relationships or omitting critical dependencies. Addressing this requires techniques that integrate \textbf{proxy variables}, infer \textbf{latent structures}, and jointly model observed and unobserved influences. Promising strategies include combining observational time-series data with limited interventional signals (e.g., actuator overrides) and leveraging \textbf{robustness-aware causal discovery} frameworks to quantify sensitivity to hidden confounding~\cite{beckers2021causal, weinberg2024causality}.

\vspace{-8pt}
\subsection{Robustness to Assumption Violations in Dynamic Environments}

Classical causal inference relies on strong assumptions such as \textit{acyclicity}, \textit{causal faithfulness}, and \textit{stationarity}. Yet, these assumptions are often violated in real-world cyber-physical systems, where concept drift, adversarial perturbations, and evolving control policies cause structural instability. For example, feedback loops across spatial control zones can violate acyclicity, and transient covariations during attacks may break faithfulness. Ensuring robustness requires future causal models to (i) incorporate \textbf{temporal priors} that distinguish persistent from transient causal links, (ii) hybridize \textbf{symbolic reasoning with deep neural models} for interpretability and flexibility, and (iii) undergo systematic \textbf{stress-testing} under adversarial and nonstationary conditions. Recent advances in \textbf{intervention-aware nonlinear causal discovery} offer a path forward, enabling robust causal reasoning even in structurally unstable or partially observed environments~\cite{li2024nonlinear}.

\vspace{-8pt}
\subsection{Grounding Counterfactual Reasoning}
While invaluable for post-mortem analysis and planning, counterfactual claims ("what if?") are difficult to validate. We advocate for grounding them by building causal simulation environments, mining implicit interventional knowledge from operational logs, and creating human-in-the-loop systems for expert feedback. When grounded in physical context and expert validation, counterfactuals can shift anomaly detection from reactive alerts to proactive, explanatory decision support~\cite{loftus2024position}. 

\vspace{-5pt}
\section{Future Directions: Towards Adaptive, Explainable, and Scalable Systems}


To advance causality-driven anomaly detection, future research must move toward integrated, adaptive, and interpretable frameworks. We outline three key directions that can bridge current limitations and enable real-world deployment.

\noindent\emph{\underline{Multi-Modality Causal Methods}:}
Current methods often rely on time-series data alone, overlooking the rich semantics available across other sources like system logs, user behavior, and configurations. Multi-modality causal frameworks can align diverse inputs—structured and unstructured—to improve anomaly detection.
For example, causal links inferred jointly from sensor data and log events can uncover propagation patterns that are invisible in any single modality. Graph-based alignment and multimodal transformers can encode such heterogeneity while preserving causal semantics, enabling fine-grained anomaly localization.

\noindent\emph{\underline{Generative AI for Causal Discovery}:}
Recent advances in generative models provide new tools for causal representation learning. LLMs can mine cause-effect pairs from domain-specific text (e.g., incident reports), while diffusion and adversarial models can simulate rare attack scenarios, augmenting data for low-frequency events.
These techniques reduce dependence on domain assumptions, support counterfactual training, and help synthesize diverse anomalous behaviors, leading to more resilient detection under novel threats.

\noindent\emph{\underline{Scalability and Adaptability}:}
Causal frameworks must evolve in tandem with the dynamic environments of modern CPS. Incremental learning strategies under streaming settings allow real-time updates to causal graphs, adapting to concept drift without retraining from scratch.
Techniques such as sparse view distillation, temporal-invariant decomposition, and expert-guided priors can balance efficiency with fidelity. Hybrid models that integrate neural encoders with causal layers are key to achieving both scalability and interpretability in large-scale deployments.

\vspace{-5pt}
\section{Conclusion}

As spatio-temporal infrastructures grow increasingly complex and mission-critical, the reliance on black-box anomaly detectors poses a fundamental threat to both reliability and security. This vision paper advocates a decisive shift toward causality-driven anomaly detection—one that goes beyond pattern recognition to uncover the underlying mechanisms that govern system behavior.
By exploring causal graph profiling, multi-view fusion, and continual causal learning, we demonstrate that causal graph models are uniquely positioned to deliver interpretable, adaptive, and robust anomaly detection in cyber-physical systems. These models not only detect anomalous behavior but also enable fine-grained diagnosis, localization, and intervention capabilities that are essential for securing spatially distributed infrastructures.
The road ahead calls for a unified research effort that advances \textbf{scalable}, \textbf{multi-modal}, and \textbf{adaptive causal frameworks}, enriched by the generative and reasoning capabilities of modern AI. We urge the community to pivot from opaque accuracy benchmarks to frameworks that prioritize transparency, intervention-readiness, and real-world applicability.
In doing so, we lay the foundation for a new generation of secure, interpretable, and spatially grounded anomaly detection systems for the infrastructures that matter most.

\bibliographystyle{ACM-Reference-Format}
\bibliography{main}


\end{document}